\documentclass[conference]{IEEEtran}
\IEEEoverridecommandlockouts
\usepackage{cite}
\usepackage{amsmath,amssymb,amsfonts}
\usepackage{algorithmic}
\usepackage{graphicx}
\usepackage{textcomp}
\usepackage{xcolor}
\def\BibTeX{{\rm B\kern-.05em{\sc i\kern-.025em b}\kern-.08em
    T\kern-.1667em\lower.7ex\hbox{E}\kern-.125emX}}

\usepackage{times}
\usepackage{epsfig}
\usepackage{graphicx}
\usepackage{amsmath}
\usepackage{amssymb}
\usepackage{url}
\usepackage{todonotes}
\usepackage{xcolor}

\usepackage{booktabs}
\newcommand{\ra}[1]{\renewcommand{\arraystretch}{#1}}
\usepackage{caption}

\usepackage{multirow}
\usepackage{siunitx}
\usepackage[english]{babel}
\usepackage{fixltx2e}
\usepackage{setspace}
\usepackage{array}
\usepackage{tabularx}
\usepackage{calc}
\usepackage{makecell}

\definecolor{gg}{RGB}{15,125,15}
\definecolor{rr}{RGB}{190,45,45}

\begin{document}

\title{Robust Handwriting Recognition \\ with Limited and Noisy Data
}


\author{\IEEEauthorblockN{
\normalsize
Hai Pham$^{\dagger}$\thanks{Corresponding author: htpham@cs.cmu.edu}, 
Amrith Setlur$^{\dagger}$, 
Saket Dingliwal$^{\dagger}$, 
Tzu-Hsiang Lin$^{\dagger}$,
Barnab\'{a}s P\'{o}czos$^{\dagger}$ 
}
\IEEEauthorblockA{
\normalsize
Kang Huang$^{\clubsuit}$, 
Zhuo Li$^{\clubsuit}$, 
Jae Lim$^{\clubsuit}$,
Collin McCormack$^{\clubsuit}$, 
Tam Vu$^{\clubsuit}$
}
\IEEEauthorblockA{
$^{\dagger}$\textit{Carnegie Mellon University}
$^{\clubsuit}$\textit{The Boeing Company}
}
}

\maketitle

\begin{abstract}
   Despite the advent of deep learning in computer vision, the general handwriting recognition problem is far from solved. 
   Most existing approaches focus on handwriting datasets that have clearly written text and carefully segmented labels.
   In this paper, we instead focus on learning handwritten characters from  maintenance logs, a constrained setting where data is very limited and noisy.
   We break the problem into two consecutive stages of word segmentation and word recognition respectively, and utilize data augmentation techniques to train both stages. Extensive comparisons with popular  baselines for scene-text detection and word recognition show that our system achieves a lower error rate and is more suited to handle noisy and difficult documents.
\end{abstract}


\section{Introduction}

Offline handwriting recognition (HWR) is a fundamental problem in computer vision~\cite{survey2000}. Unlike online HWR where a stroke direction is a valuable cue \cite{hmm2014warp, graves2009ctc},
in the offline setting, we simply have access to an image of the final handwritten words instead. 
Nowadays, although data can be easily digitized and stored, there is still a need to recognize and digitize handwritten paper documents~\cite{blikstein2016multimodal,likforman2007text}.

Despite the significant demand, there are few efficient methods  able to tackle this problem due to 
the difficulty of designing a holistic solution suitable across various forms of input.
The first challenge is to segment forms (\textit{i.e.} images containing lines) properly to facilitate the recognition process. The most common method is to use a heuristics line-level segmentation \cite{graves2009ctc,graves2007ctc}. However, this is often impractical since words and characters are not usually handwritten along straight lines. The second challenge is to build a model capable of recognizing and generalizing diverse handwriting styles.
Furthermore, in some resource-constrained settings where we have limited access to real data, it is infeasible to manually build large-scale handwriting recognition datasets such as IAM-DB \cite{marti2002iam}, or SD19 \cite{sd19}.
It therefore becomes necessary to find a powerful solution that does not require a large quantity of real data. This problem is ubiquitous in practice, 
in that we only have access to limited data with inherent noise. 
Typically in such settings, people rely on commercial systems which are prohibitively expensive, or open APIs such as Google Cloud Vision\footnote{\url{https://cloud.google.com/vision/}} or Tesseract \cite{tesseract} which typically perform poorly as they are mainly designed for \textit{printed} text and for dealing with many languages with a single model.

\begin{figure}[t!]
\begin{center}
\includegraphics[width=0.8\linewidth]{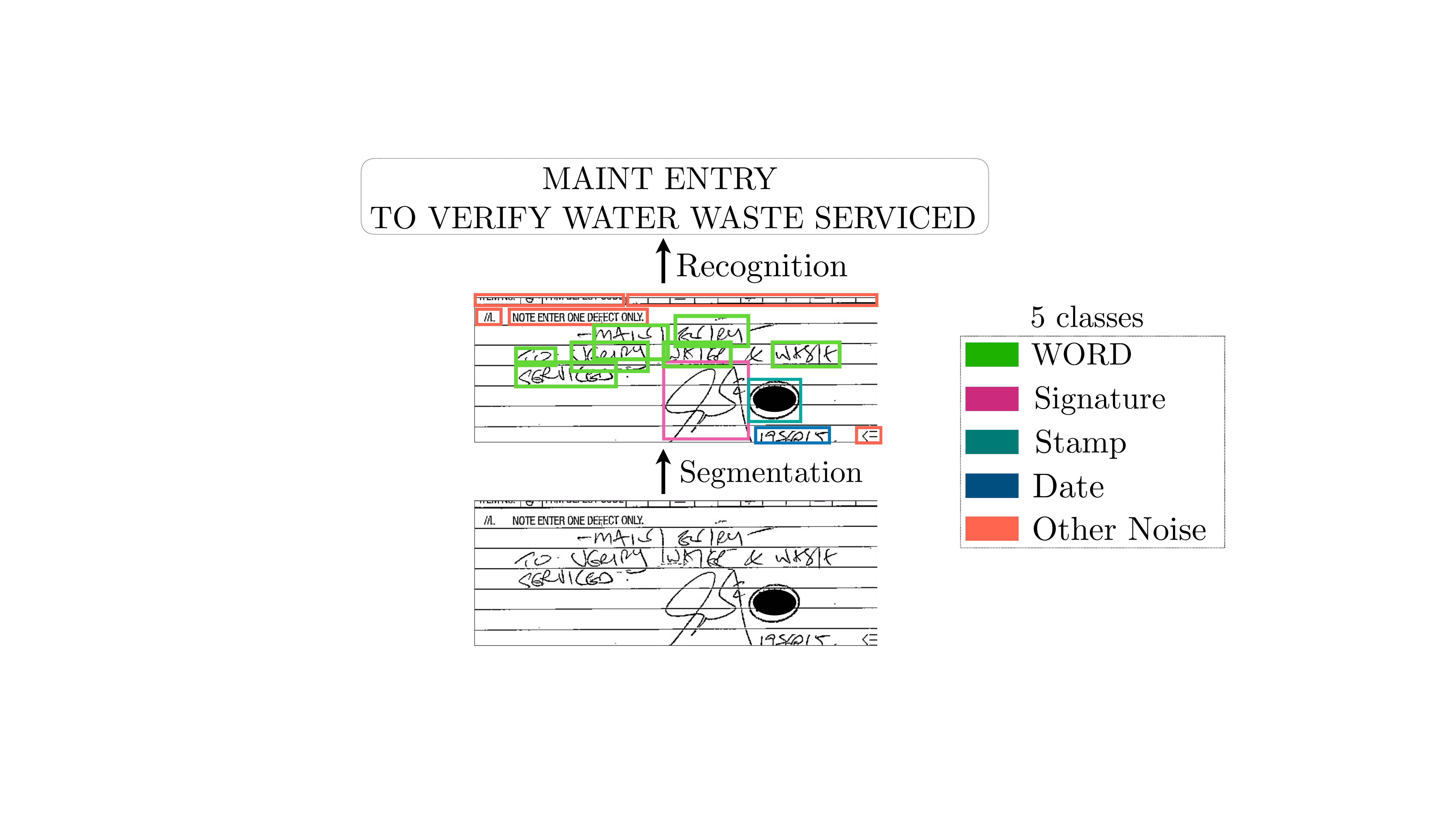}
\end{center}
    \caption{Our model can handle noisy forms by localizing unaligned texts, filter out other types of noise to recognize the sentence(s) in the correct order of words. We hide stamps' contents for security reason.
    \vspace{-2mm}
    }
\label{fig:overview}
\end{figure}

\begin{figure}[th!]
\begin{center}
\includegraphics[width=0.75\linewidth]{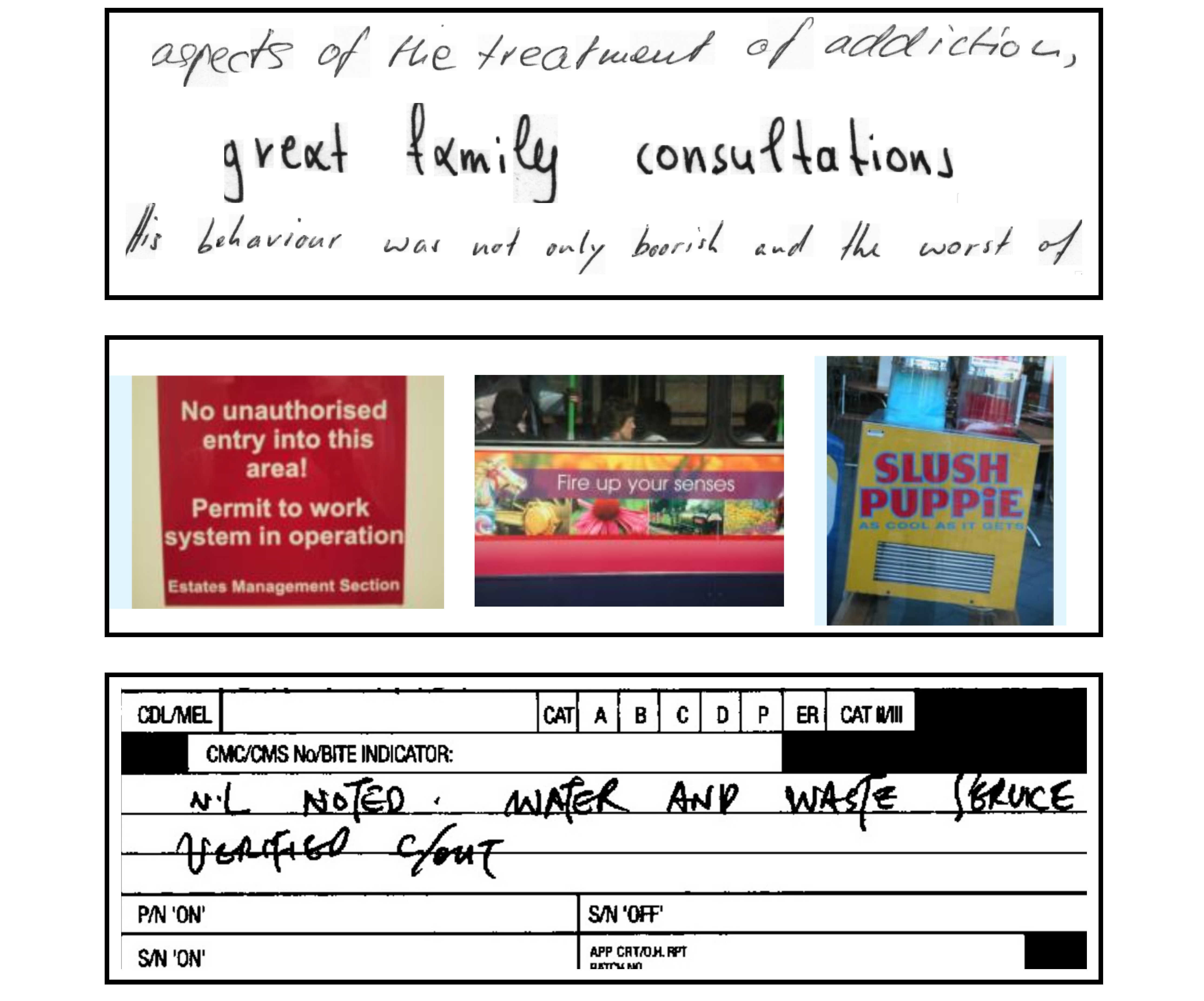}
\end{center}
   \caption{Some samples from 3 different problems. \textit{Top}: three lines of IAM~\cite{marti2002iam} dataset which has handwritten text on blank background; most solutions segment them into lines without clarifying segmentation quality. \textit{Middle}: ICDAR~\cite{karatzas2015icdar} dataset for scene-text recognition which has printed text with random background. \textit{Bottom}: our BHD dataset which combines the difficulties of the other two: multi-style, unaligned handwritten text in the whole form (not lines) and noisy background.
   \vspace{-4mm}
   }
\label{fig:samples}
\end{figure}

\vspace{-1mm}
Furthermore, our dataset is much more difficult than IAM-DB or SD19. First, it has limited and noisy data and annotation. Second, it combines the difficulties of the classical HWR datasets and scene-text detection and recognition ones (Fig.~\ref{fig:overview} and~\ref{fig:samples}). As a result, we modularize the problem into two stages in order to make it more tractable to train two separate deep models. 
In the first stage we employ a object detection model, such as R-FCN \cite{dai2016r}, to detect words from the background with various types of noise.
The resulting segments are fed into a recognition model in the second stage which can be a word-based or a character-based model. 


We evaluate performance based on multiple metrics and show  
that our system is able to detect words in challenging settings with high accuracy. In addition, we also demonstrate advantages over several state-of-the-art (SoTA) methods for the related tasks of scene-text detection and recognition.
To sum up, our contributions are as follows: we show (i) that in a constrained setting as defined above, a two-phase approach including segmentation and recognition at the form level (instead of line level) is an efficient method, 
and (ii) extensive experiments and analysis that provide guidelines for similar applications in this setting.

\section{Related Work}

For offline HWR, there have been many achievements using the classical HMM-based models \cite{hmm2007,lexiconfree,hmm2014warp,hmm2000}. 
Later, with the advent of deep learning, Recurrent Neural Network based approaches, such as using LSTM \cite{lstm}, gained new successes in this setting \cite{voigtlaender2015sequence,puigcerver2017multidimensional ,kang2018convolve}. Following this line, there have been also some other solutions that also employ convolutional neural network (CNN) such as in \cite{dutta2018improving}, and using CNN plus language-based features \cite{poznanski2016cnn, krishnan2016deep}.  However, in comparison to their settings, our variable-sized forms are more challenging for they include horizontal lines running across the document which contribute to noise since the text doesn't necessarily conform to these lines. Furthermore, the content is mixed with other random noise such as signatures, stamps or other unrecognized marks caused by scanners or inks. 
%
%
Given such difficult inputs, our model can directly process whole forms properly, 
in contrast to these existing solutions that rely on heuristic methods for line-level segmentation. 



A closely related problem to our method is segmentation for which there have been some heuristic \cite{survey2000} or HMM-based \cite{automaticsegment} methods. 
Our model instead relies on deep segmentation frameworks which are usually employed for object detection tasks~\cite{dai2016r,lin2018focal,yolov3,maskrnn,fasterrcnn}.
Unlike those methods that learn
to predict a regression bounding box and detect an object at the same time, in our segmentation phase, we reduce the task to a more tractable problem of only predicting a bounding box covering a word, and leave the recognition job to a downstream task. We retain the order of the words while doing this so as to ensure that sentence or document level meaning is retained.



Finally, the most related approaches to our model are those designed for scene-text detection and recognition \cite{borisyuk2018rosetta,zhou2017east,deng2018pixellink,baek2019character,luo2019moran} although their problem is different from HWR. 
But unlike their solutions which deal mainly with printed text, our setting is intrinsically harder due to the inherent difficulty in recognizing handwritten text of different styles.

\begin{figure}[t!]
\begin{center}
\includegraphics[width=\linewidth]{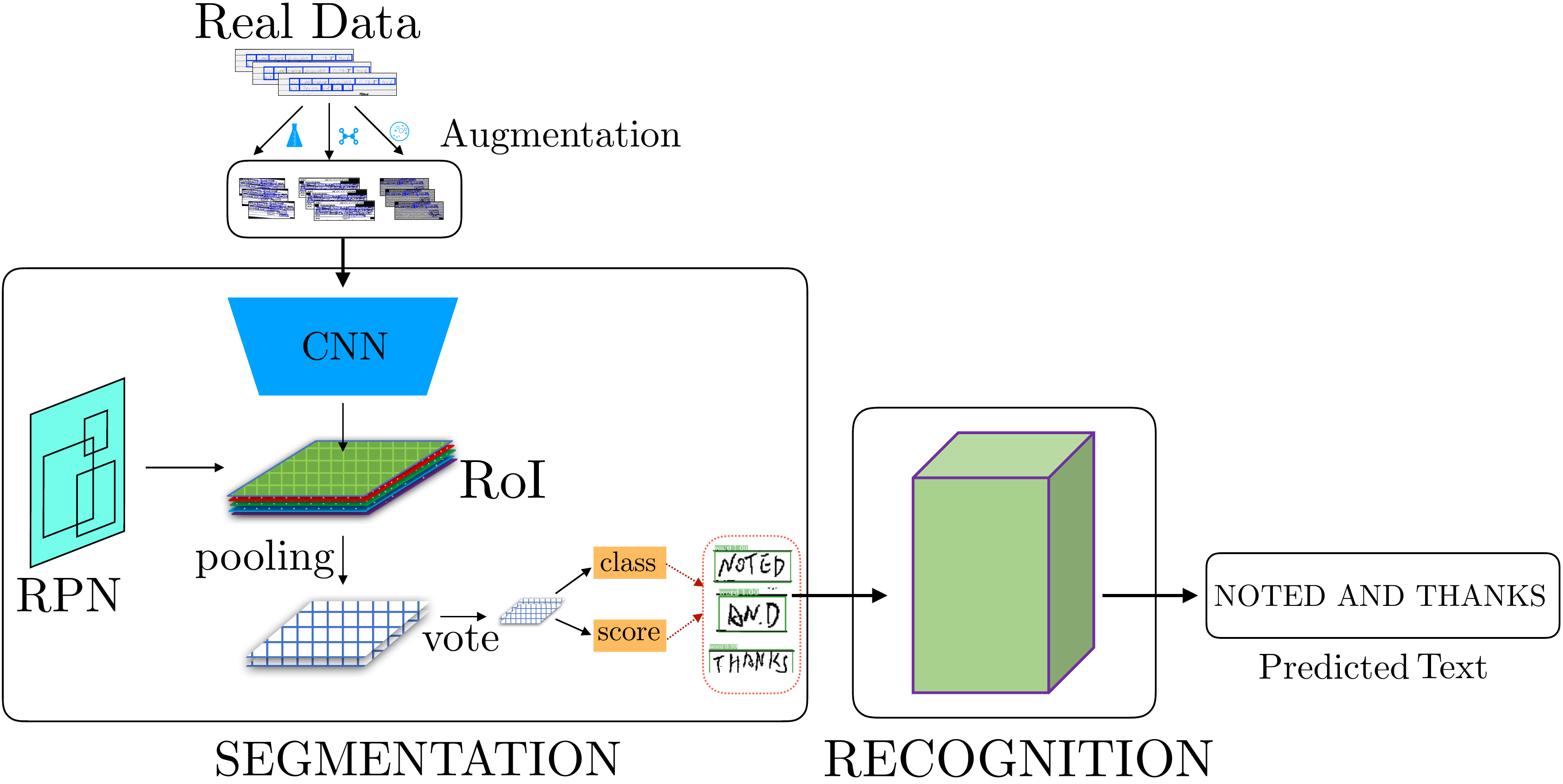}
\end{center}
   \caption{Our model uses data augmentation to train a segmentation module (locating words amidst background noise) and word recognition (word or character-based) models.
  \vspace{-4mm}
   }
\label{fig:architecture}
\end{figure}
\vspace{-2mm}
\section{Model} \label{sec:model}
Our inputs are rectangular images of varying sizes containing handwritten sentences, often in unaligned lines and with lots of noise and other irrelevant content such as stamps, signatures and other types of random noise. 
Our goal is to recognize those relevant sentences, and output the corresponding texts for further data analysis purposes. 

\subsection{Choice of Two-phase Model} 

As mentioned above, we design a two-phase approach (Fig.~\ref{fig:architecture}): segment the entire form into words (in the presence of noisy content) while maintaining their original order, and recognize each word individually. There are many reasons for this approach. First, we have very few annotated samples (Table~\ref{tbl:data}), thus the generalizability of our model is benefited from the inductive bias of the two stage approach. Second, the difficulties of the forms are unusual. Due to unaligned texts, it is impossible to segment forms into lines without affecting the content as in other HWR methods. Furthermore, like scene-text recognition datasets, our forms have many types of noise (Fig.~\ref{fig:overview} and~\ref{fig:samples}). Third, this approach is interpretable and easier to train and debug. Finally, it becomes easier to perform parallel training of the two stages across limited resources, allowing for better quality control and modularity in design.





\subsection{Word Segmentation} \label{sec:word_segmentation}

Instead of trying to predict the correct bounding boxes and recognize the words inside simultaneously, 
the word segmentation phase only focuses on drawing correct bounding boxes at the word level, and leaves the recognition job as a downstream task. We choose this design for the following reasons. First, word-level segmentation is used since  separating spaces among words (as opposed to characters) is much more feasible in practice (especially in cursive handwriting). Second, as explained previously, line-level segmentation is not preferred since in our setting words are often not aligned horizontally. 

In terms of architecture, since HWR is different from object detection where detection is only a proxy, we explore multiple options like R-FCN \cite{dai2016r}, Faster R-CNN \cite{fastrcnn} and YOLO-v3 \cite{yolov3} to identify which kind of architecture is most suited for our HWR pipeline. 
Although the core components of those detection methods remain unchanged, it is worth noting two important changes in adapting such methods. 
First, given word segmentation is an intermediate step, we simplify this phase by limiting the number of classes to only 5 (Fig. \ref{fig:overview}), with the main goal being extracting text out of the forms without having to recognize its content. 
Second, based on the nature of our dataset, we change the segmentation input to grayscale images with only 1 channel. As a result of these two adjustments, our segmentation phase is much easier and faster to train compared with their original uses.

\subsection{Word Recognition}
\label{sec:word_recognition}

For each form, this module takes the bounding boxes (as images) from the Word Segmentation module as inputs, and outputs a word for each bounding box. 
Based on the coordinates given by the Word Segmentation module, we are able to reconstruct the entire sentence from individual words.
And because the complications of the input forms, we experiment with 3 different models namely Word Model, Character Model and CTCSeq2Seq Model, as detailed below. 

\subsubsection{Word Model}
\label{susubsec:word}

The word model is a CNN-based image classification network which uses an augmented Resnet-18~\cite{he2016deep} to predict words from a predefined word vocabulary.   
Furthermore, due to the low resolution of our input images, we adjust Resnet-18 to only have a stride size of 1 instead of 2 in the residual blocks.  This model is simple, but is only capable of predicting words within the predefined vocabulary of 998 words. 

\subsubsection{Character Model}
\label{susubsec:char}

This model shares its architecture with the Word Model, which enables the benifit of initializing weights from a pretrained Word Model, 
except that it uses a CTC loss~\cite{graves2007ctc,graves2009ctc,voigtlaender2016handwriting} instead of cross-entropy loss. For this reason, it predicts a sequence of characters instead of a single word at a time.
Furthermore, the last fully-connected layer in Resnet-18 is replaced with a convolutional layer to reshape the output from  
$H * W * D$ to $1 * W/2 * C$, where $C$ is the cardinality of the character prediction space. 

By using CTC, this model has two advantages over Word Model.   
First, CTC largely reduces the prediction space from 998 words to 35 alpha-numeric characters (our dataset does not have the letter ``Z"), making it agnostic to word vocabulary size. Second, it enables the model to predict unseen words.

\begin{figure}[t!]
\begin{center}
\includegraphics[width=\linewidth]{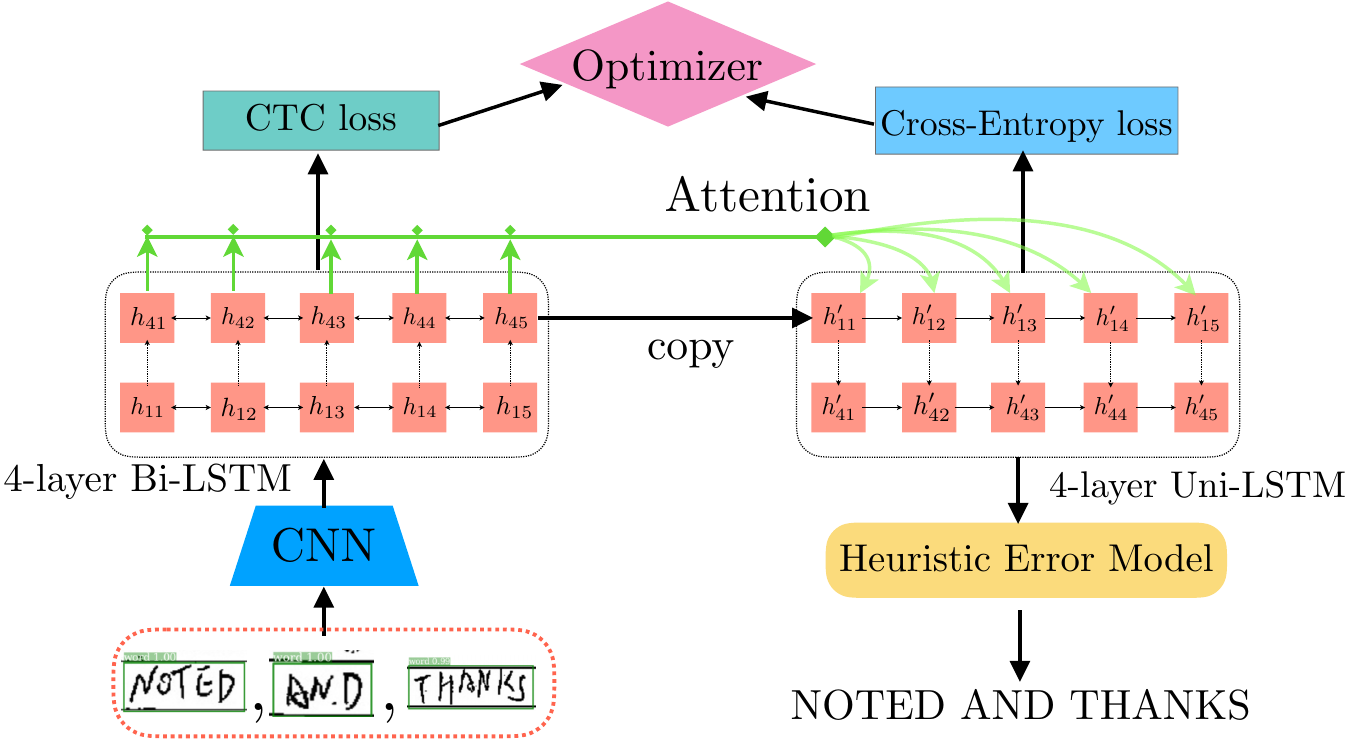}
\end{center}
    \caption{Our CTCSeq2seq model contains 3 core modules: Feature Extraction which is CNN-based, Encoder and Decoder that combined form a Seq2Seq model. The encoder uses \textit{CTC loss} which helps with alignments of the frames to the outputs. 
    \vspace{-6mm}
    }
\label{fig:hybrid_model_arch}
\end{figure}

\subsubsection{CTCSeq2Seq Model} \label{susubsec:hybrid}
Our motivation for this model is to learn the embedded latent representation of images that can be decoded into text. 
As shown in Fig. \ref{fig:hybrid_model_arch}, the model can be broken down into 3 main blocks: Feature Extraction, Encoder and Decoder.
The model loss is the weighted sum of CTC loss (Encoder) and softmax cross-entropy loss (Decoder). Except for those 3 main modules, there is an edit-distance based error module which corrects a predicted out-of-vocabulary word within a maximum of 2 wrong characters compared to a known word. 

\textbf{Feature Extraction:} 
This module accepts variable-sized input images, each of which has a single word. 
It firstly resizes inputs to the same height but not necessarily to the same width. Next, it slices each one into small patches of equal widths (as illustrated in Fig.~\ref{fig:attention_map}). 
Finally, it extracts CNN-based features out of the patches using a custom VGG \cite{vgg}. 

\textbf{Encoder:} For our encoder,  we use a 4-layer bidirectional LSTM 
that takes inputs from the Feature Extraction module. Since each input word is segmented into many sequential equal-height patches, the LSTM can model their relations into a hidden representation. Another key feature of this module is to have a CTC loss to enforce reconstruction of the original characters, so that the embedded representation is learned effectively. 

\textbf{Decoder:} This module is a 4-layer unidirectional LSTM 
that consumes the hidden representation from the Encoder and has an  
attention module \cite{luong2015effective} which calculates the weighted average of each output with the entire input sequence. 
This mechanism helps the model learn to focus on more important patches. 

In addition, this module uses the softmax cross-entropy loss normalized by the length of input, since we have variable-length sequences of patches. 
Finally, it also predicts among 35 alpha-numeric characters, same as Character Model (Section \ref{susubsec:char}) which also ignores punctuation in the datasets. 


\section{Experiments} 

\begin{table}[ht!]
\fontsize{8.5}{10}\selectfont
\setlength\tabcolsep{7.0pt}
\centering
\begin{tabular}{l | c | c c c}
\Xhline{3\arrayrulewidth}
Dataset & Type & Train & Valid & Test\\
\hline \hline
\multirow{2}{*}{Segmentation} 
    & Real         &   2,358    &    -    &   1,362     \\
    & +DA          &  40,159    &    -    &    -       \\ 
\Xhline{0.5\arrayrulewidth}
\multirow{2}{*}{Recognition} 
    & Real       &     6,639    &  3,400  & 1,249  \\
    & +DA        &   660,000    &    -    &   -    \\ 
\Xhline{0.5\arrayrulewidth}
Pipeline         &  Real        &    -    &   - &  1,362   \\
\Xhline{3\arrayrulewidth}
\end{tabular}
\caption{Statistics of BHD dataset. We have 2 types of data (i) Real and (ii) +DA : real images with data augmentation. For each model, we only have a single test set from real forms, and the one used for Pipeline evaluation is shared with Segmentation. Data augmentation is a key preprocessing step to get more samples and styles for training deep models.\vspace{-4mm}}
\label{tbl:data}
\end{table}

\subsection{Dataset} \label{subsec:data}


Our in-house BHD dataset, as shown in Table \ref{tbl:data}, comprises of maintenance logbooks in which there are
many aerospace terms or abbreviations that do not appear in the normal English vocabulary.
Each image is grayscale and may contain from 3 to 50 bounding boxes. Moreover, in addition to the presence of unusual aerospace terms, there are many arbitrary part numbers (\textit{e.g.}, ``W308003-12239-22''). 
As mentioned earlier, our forms contain multiple horizontal lines, with signatures, stamps, dates and other types of noise, making our task even more challenging. 
Finally, to create word vocabulary, we use \texttt{tf-idf} to retrieve the first 1000 words from digitized maintenance logbooks, then remove 2 outliers to finally have 998 words. 

Furthermore, our manual inspection of the BHD dataset reveals that 
in several cases the strokes from adjacent words are connected to each other, while in other cases, the characters in a word are quite far apart, which tempts any object detection model to confuse multiple words with just one. This makes BHD more challenging than ICDAR and other scene-text detection and recognition datasets. 

\subsection{Training Data Augmentation} \label{subsec:data_augmentation}

Because we have limited data, and our model contains deep neural networks that are typically data hungry, data augmentation is an important technique to increase the effective size of BHD prior to training, and to improve the generalization capability of our models. 
In particular, we use two data augmentation techniques for 
both segmentation and recognition tasks. 
First, we use several types of noise including pepper, stroke and Gaussian noises. Second, we employ local image transformations that are erosion, dilation and flipping. 






\subsection{Evaluation Metrics}
\label{subsec:metrics}
\textbf{Segmentation:} 
We use the canonical MaP metric \cite{everingham2010pascal} to evaluate segmentation performance against our annotation in the BHP real-form test dataset. 

\textbf{Recognition:} 
We use word accuracy (WA) and Character Error Rate (CER) to evaluate our recognition models. While WA simply calculates the average number of predicted words that exactly matches with ground truths, CER is calculated as
$
    \text{CER} = (\text{D}(w_{gt}, w_{predict}) \times 100) \, / \, |{w}_{gt}| \, (\%),
$
where $\text{D}(w_{gt}, w_{predict})$ is the minimum Damerau-Levenshtein edit distance~\cite{damerau1964technique} between the ground-truth word $w_{gt}$ and predicted word $w_{predict}$, and $|w_{gt}|$ is the number of characters in $w_{gt}$. 

\textbf{Full Pipeline:} Our pipeline takes a form as input, and outputs a sequence of predicted words. Therefore we use Word Error Rate (WER) and CER to evaluate performances. For WER, we treat every word as a character. For CER, we concatenate the sequence of words by inserting a space between every two words and treating the concatenated sequence as the predicted string.

\subsection{Baselines}
\label{sec:baselines}


Since different models require different sets of annotations (\textit{e.g.} many HWR models expect noise free input), we cannot fairly compare our full pipeline performances with many SoTA methods for HWR. As a result, the only close HWR pipeline we compare our model with is Convolve-Attend-Spell~\cite{kang2018convolve} (after it is fine-tuned on the full-pipeline dataset) which has the capability of accepting the entire form as an input and to some extent is also robust to noise. 


However, we can compare each phase of our pipeline with segmentation and recognition baselines developed for scene-text detection. 
For segmentation, we use EAST \cite{zhou2017east}, PixelLink\cite{deng2018pixellink} and CRAFT \cite{baek2019character}. 
In order to have a fair comparison, we fine tune EAST and PixelLink\footnote{The same cannot be done for CRAFT due to its code's unavailability.} (trained on ICDAR 2015 \cite{karatzas2015icdar}) and only compare on the \textit{word} class, which is the ultimate goal. 
For recognition, we use MORAN \cite{luo2019moran} which is pre-trained on synthetic images \cite{jaderberg2014synthetic,gupta2016synthetic} and subsequently fine tuned on BHD recognition training data. 
And last, for full pipeline, we combine PixelLink and MORAN, for which the full training codes are available.

\section{Results and Discussion}

We compare the performances of our approach to the baselines for the full pipeline, segmentation and recognition. We also perform an ablation study on the impact of segmentation on the full pipeline.  

\begin{table}[h!]
\fontsize{8}{10}\selectfont
\setlength\tabcolsep{7.0pt}
\centering
\begin{tabular}{l | l | c c}
\Xhline{3\arrayrulewidth}
Segmentation   & Recognition & WER($\downarrow$) & CER($\downarrow$)\\
\hline \hline
\multirow{2}{*}{R-FCN~\cite{dai2016r}}    & Word  & 31.5 & 22.9 \\
                           & CTCSeq2Seq  & \textbf{30.1} & \textbf{18.5} \\
\hline 
PixelLink~\cite{deng2018pixellink} & MORAN~\cite{luo2019moran}    & 80.7        & 47.4 \\ 
\hline 
\multicolumn{2}{c}{Convolve-Attend-Spell~\cite{kang2018convolve}} & 38.9 & 24.1 \\
\Xhline{3\arrayrulewidth}
\end{tabular}
\caption{Full pipeline performance of our best model compared to the baselines. Our model significantly outperforms all the baselines in both WER and CER metrics. \vspace{-4mm}
}
\label{tbl:best_pipeline_results}
\end{table}

\subsection{Full Pipeline Results}
\label{subsec:full_pipe}

The full pipeline results are shown in Table \ref{tbl:best_pipeline_results}. We observe that that R-FCN \cite{dai2016r} in conjunction with CTCSeq2Seq (both of which are trained on the +DA dataset) yields the best performance, and significantly outperforms the baseline models.

Furthermore, Fig. \ref{fig:full_results} illustrates some qualitative results. The R-FCN is able to filter out several types of noise in each form and pick out the correct bounding boxes with almost 100\% confidence for all words. Furthermore, our CTCSeq2Seq is able to detect words and characters of various styles, orientations and intensities. However, the baseline one makes lots of mistakes in word localization, which are compounded in the second phase of recognition. 


\begin{table}[h!]
\fontsize{8}{10}\selectfont
\setlength\tabcolsep{3.0pt}
\centering
\begin{tabular}{c | c | c | c | c | c | c}
\Xhline{3\arrayrulewidth}
& EAST & CRAFT & PixelLink & R-FCN & Faster-RCNN & YOLO-v3 \\
\hline \hline
AP  & 38.9 & 12.8 & 81.6 & 89.0 & \textbf{89.1} & 86.0\\
\Xhline{3\arrayrulewidth}
\end{tabular}
\caption{AP score comparison on the \textit{word} class (IoU=50\%). Our three models significantly outperform the baselines.
\vspace{-4mm}
}
\label{tbl:seg_baselines}
\end{table}

\subsection{Segmentation Results}
\label{subsec:seg_results}

\begin{table*}[th!]
\setlength{\tabcolsep}{18pt}
\ra{1.2}
\centering
\begin{tabular}{cc}
\includegraphics[height=1.9cm]{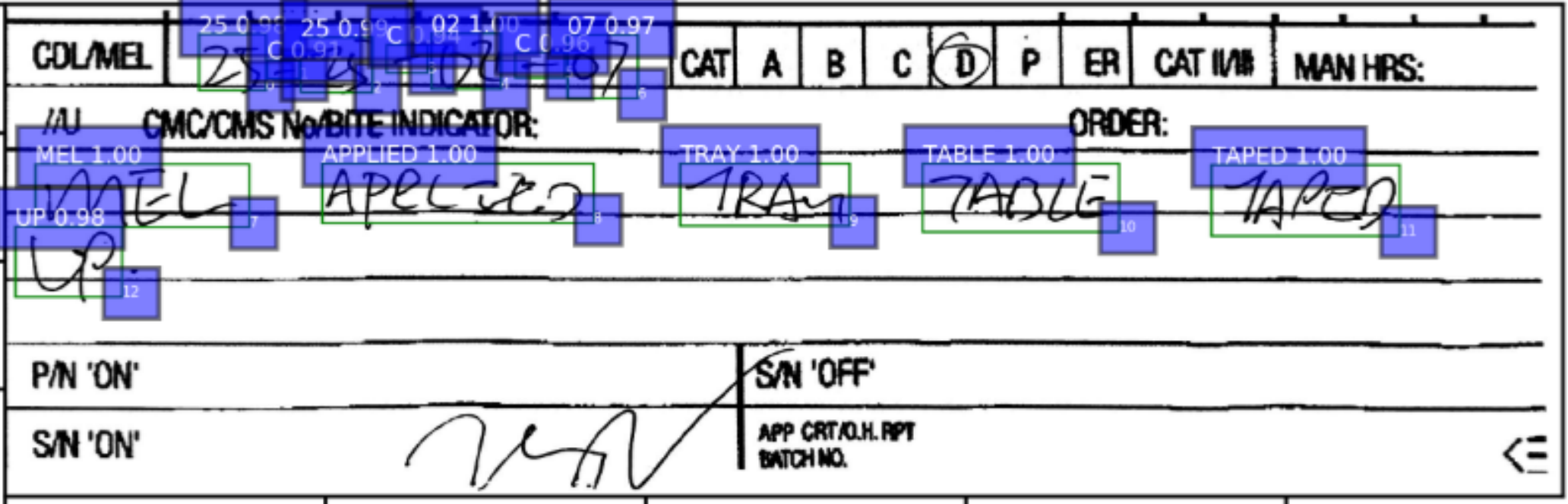} &
\includegraphics[height=1.9cm]{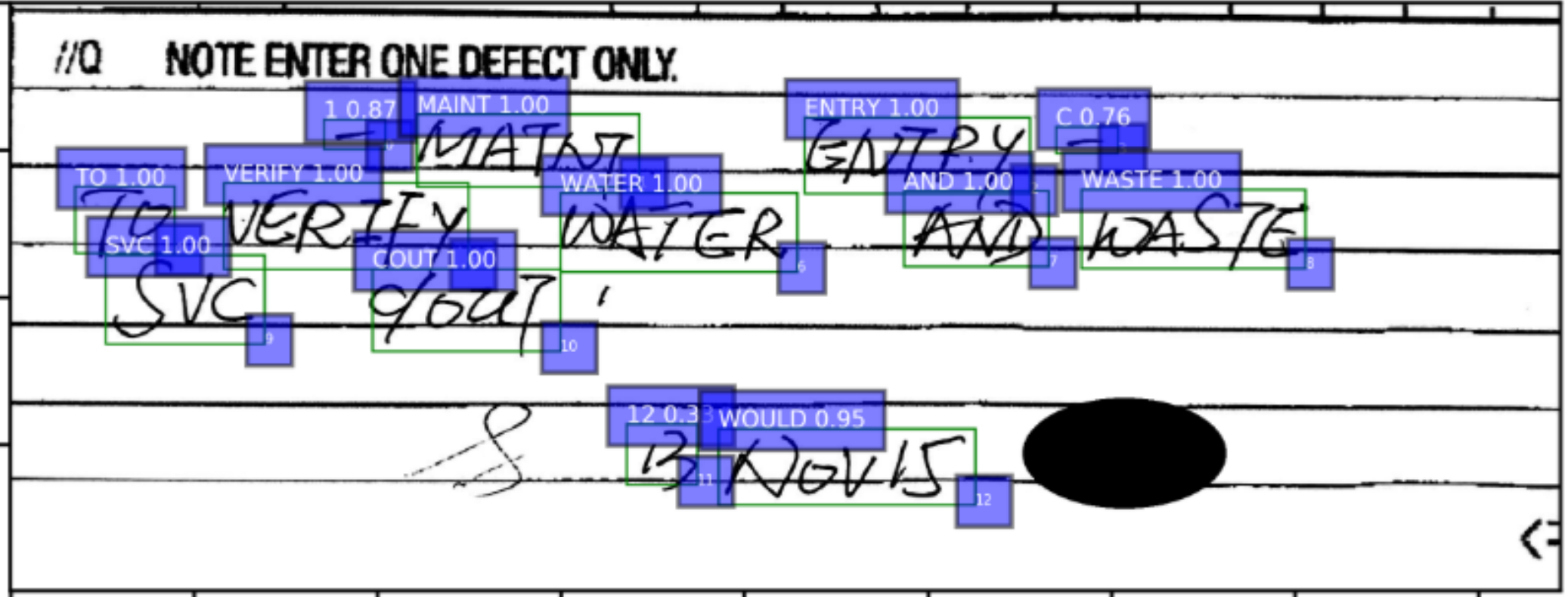} \\
\includegraphics[height=1.9cm]{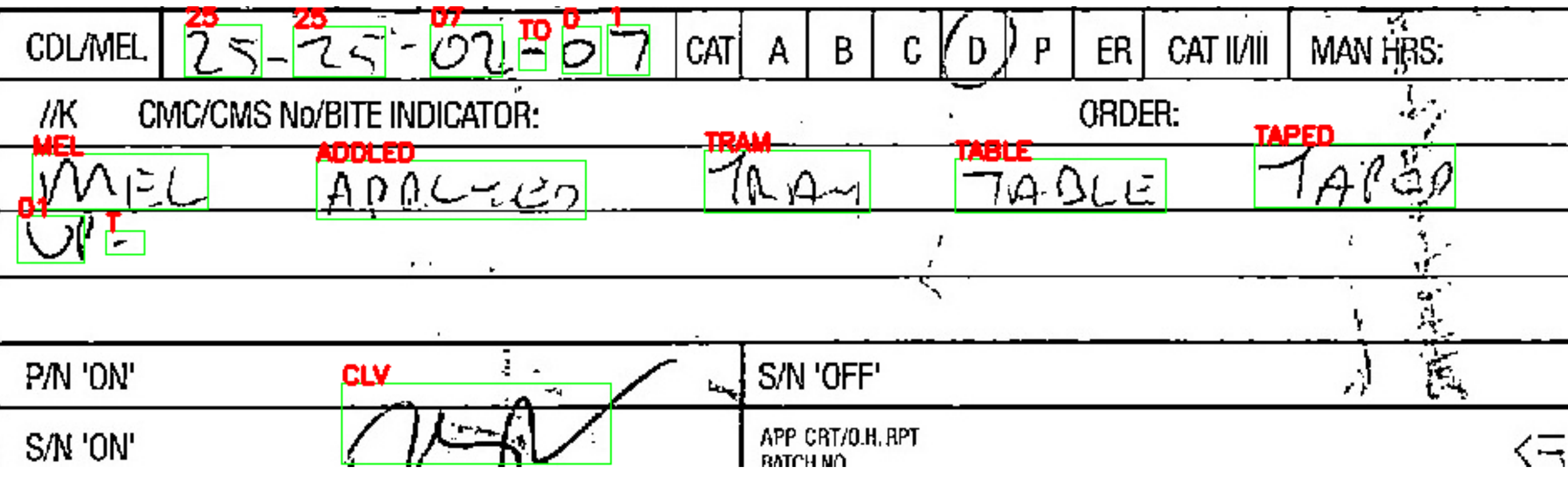} &
\includegraphics[height=1.9cm]{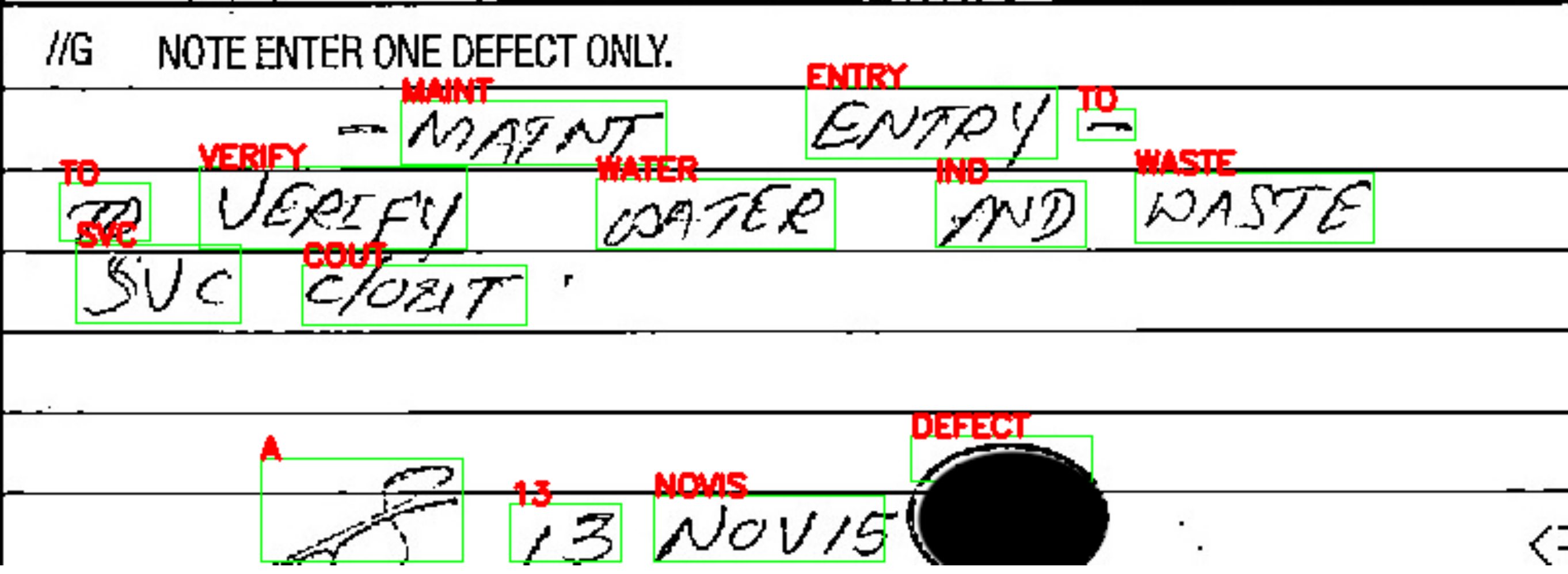} \\
\end{tabular}
\captionof{figure}[]{Full pipeline qualitative results of our model R-FCN~\cite{dai2016r} + CTCSeq2Seq (top) and the baseline PixelLink~\cite{deng2018pixellink} + MORAN~\cite{luo2019moran} (bottom). Ours performs much better in both locating words and recognizing them.\vspace{-2mm}}
\label{fig:full_results}
\end{table*}

\begin{table}[t!]
\fontsize{8}{10}\selectfont
\setlength\tabcolsep{6.0pt}
\centering
\begin{tabular}{l | c | c | c | c}
\Xhline{3\arrayrulewidth}
\multirow{2}{*}{Class}  & R-FCN & R-FCN & Faster R-CNN & YOLO-v3 \\ 
                        & (Real) & (+DA) & (+DA) & (+DA) \\ 
\hline \hline
Word     &  85.8 & 89.0           & \textbf{89.1} & 86.0 \\
Signature & 67.9 & \textbf{78.2}  & 43.3 & 40.8 \\
Stamp     & 86.6 & \textbf{89.9}  & 10.7 & 84.2 \\
Date      & 70.1 & \textbf{82.9}  & 24.7 & 62.9  \\
Noise     & 18.2 & 17.4           & \textbf{27.3} & 15.2  \\
\Xhline{0.5\arrayrulewidth}
Average & 65.7 & \textbf{71.3} & 39.0 & 57.8  \\ 
\Xhline{3\arrayrulewidth}
\end{tabular}
\caption{AP scores for Segmentation models R-FCN \cite{dai2016r}, Faster R-CNN \cite{fasterrcnn} and YOLO v3 \cite{yolov3}. R-FCN significantly outperforms others on most classes with augmented training data (IoU=50\%).
}
\label{tbl:multidatasets}
\end{table}

As shown in Table \ref{tbl:seg_baselines}, our three segmentation models clearly outperforms all baseline methods, especially on EAST and CRAFT. 
While EAST fails to split large bounding boxes, leading to a low recall (18.4\%), CRAFT's pretrained model mistakes printed words for handwritten text and therefore has a low precision (21.4\%). Finally, since PixelLink is trained on BHD, it can achieve a decent score of 81.6\% AP. 

Additionally, considering only our models, Table~\ref{tbl:multidatasets} 
shows that data augmentation leads to improvements on AP for R-FCN (especially for rare categories like \textit{Signature} or \textit{Date}). R-FCN with position-based scores is particularly effective in tackling translation variance \cite{dai2016r} for handwriting recognition where the Region-of-Interest (RoI) is fairly small (as seen in Fig.  \ref{fig:full_results}). 
\subsection{Recognition Results}

\begin{table}[t!]
\fontsize{8}{10}\selectfont
\setlength\tabcolsep{7.0pt}
\centering
\begin{tabular}{l | c | c | c}
\Xhline{3\arrayrulewidth}
Model & Dataset & WA ($\uparrow$) & CER($\downarrow$)  \\
\hline \hline
\multirow{2}{*}{Word} 
    & Real        & 76.1          & 20.4 \\
    & +DA         & \textbf{96.1} & \textbf{2.6} \\
\Xhline{0.5\arrayrulewidth}
\multirow{2}{*}{Character} 
    & Real        & 5.0    & 62.8 \\
    & +DA         & 76.3    & 9.7 \\
\Xhline{0.5\arrayrulewidth}
\multirow{2}{*}{CTCSeq2Seq} 
    & Real        & 87.1 & 7.8 \\
    & +DA         & 94.9 & 3.2 \\
\Xhline{0.5\arrayrulewidth}
\multirow{2}{*}{MORAN}
    & Real        & 91.7 & 3.4 \\
    & +DA         & \textbf{96.4} & \textbf{1.5} \\
\Xhline{3\arrayrulewidth}
\end{tabular}
\caption{Comparison on recognition models (on Recognition dataset) given ground-truth bounding boxes. 
Our Word Model and MORAN \cite{luo2019moran} perform the best compared to others.\vspace{-2mm}}
\label{tbl:cls_results}
\end{table}

As demonstrated in Table \ref{tbl:cls_results}, our Word Model achieves the similar performances to the best performer MORAN in both WA and CER given ground-truth bounding boxes. 
Even being initialized with Word Model's pretrained weights, the Character Model under-performs the other two by a huge margin.   
We suspect the reason is that CTC is hard to train, and may require more training data or more complex techniques.  





\subsection{Ablation Study}
\label{sec:ablation_study}

\begin{table}[t!]
\fontsize{8}{10}\selectfont
\setlength\tabcolsep{7.0pt}
\centering
\begin{tabular}{l | c | c | c}
\Xhline{3\arrayrulewidth}
Recognition & Segmentation &  WER($\downarrow$) & CER($\downarrow$)  \\
\hline \hline
\multirow{3}{*}{Word} 
& Ground Truth     & 15.1  &  9.5  \\
& R-FCN           & 18.3  &  13.2  \\
& Faster R-CNN         & 19.1  &  21.0  \\
\Xhline{0.5\arrayrulewidth}
\multirow{3}{*}{CTCSeq2Seq} 
& Ground Truth         & \textbf{14.1}  & \textbf{8.2} \\
& R-FCN      & 18.9  &  12.3 \\
& Faster R-CNN    & 19.8  &  19.5 \\
\Xhline{0.5\arrayrulewidth}
\multirow{2}{*}{MORAN} 
& Ground Truth    & 49.2 & 25.7  \\
& PixelLink      & 80.7  &  47.4 \\
\Xhline{3\arrayrulewidth}
\end{tabular}
\caption{
Impact of different Segmentation methods on the full pipeline (on Pipeline dataset). Our models clearly outperform the baselines, and CER is much higher if we replace R-FCN by Faster R-CNN. 
\vspace{-4mm}}
\label{tbl:ablation_seg_pipline}
\end{table}

\begin{figure}[h!]
\begin{center}
\includegraphics[width=0.55\linewidth]{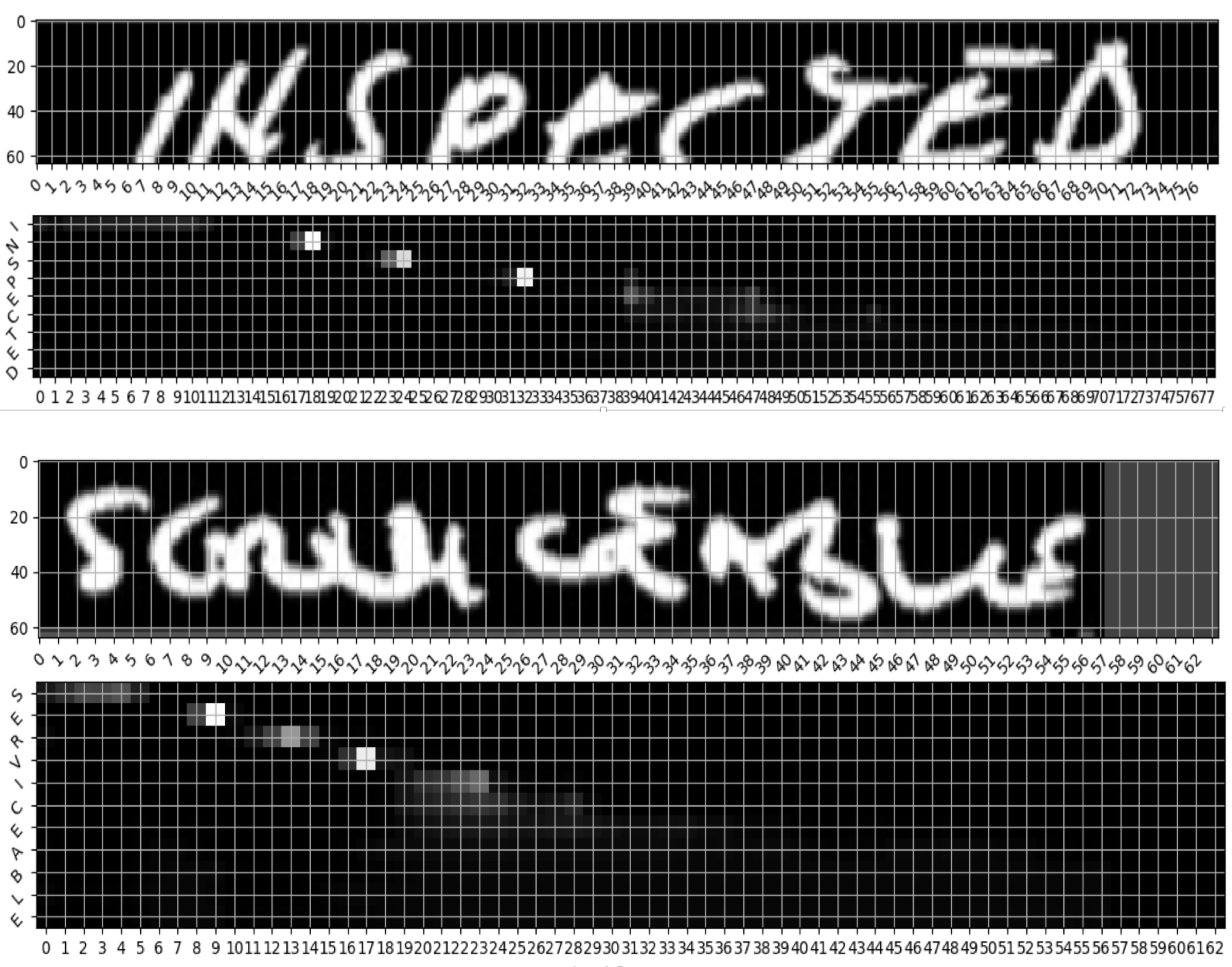}
\end{center}
    \caption{Attention map results of CTCSeq2seq model for 2 words: \texttt{INSPECTED} and \texttt{SERVICEABLE}. The upper image is raw input 
    and the lower one is the corresponding attention map.  Brighter squares indicates higher weights (focusing more in decoding). 
    After first several characters are recognized, the model can infer the rest of characters without relying on encoder information.
    \vspace{-2mm}
    }
\label{fig:attention_map}
\end{figure}

We study how different segmentation models affect pipeline performance on the same recognition model. 
As shown in Table~\ref{tbl:ablation_seg_pipline}, our models perform much better than the baselines, and CTCSeq2Seq is the best recognition model. As shown in Fig.~\ref{fig:attention_map}, CTC loss combined with attention module significantly helps with character recognition, making the CTCSeq2Seq the best choice for our full pipeline. 

And interestingly, CER increases much more than WER when replacing R-FCN with Faster R-CNN. 
Our empirical analysis reveals that R-FCN tends to give predictions with higher confidence scores and in difficult cases, it predicts more bounding boxes than Faster R-CNN in the segmentation phase.
Finally, given ground-truth bounding boxes, both WER and CER decrease but only to a limited extent. 
This suggests that the segmentation module is not the bottleneck of whole pipeline system, and we should focus more on the recognition module to increase pipeline performance.

\section{Conclusion}
In this paper, we focused on HWR for noisy and challenging maintenance logs, a previously overlooked domain in this field. We presented a two-stage approach that can process the entire forms directly without the need of segmenting them into lines. Our experimental results show that our approach significantly outperforms the HWR and scene-text detection and recognition baselines on the full pipeline while 
achieving high accuracies on the individual phases of word segmentation and recognition.









{
{
\bibliographystyle{plain}
\bibliography{refs}
}
}


\end{document}